%% file: main.tex
\documentclass{article}
\usepackage{arxiv}

\usepackage[utf8]{inputenc} 
\usepackage[T1]{fontenc}    
\usepackage{hyperref}       
\usepackage{url}            
\usepackage{booktabs}       
\usepackage{amsfonts}       
\usepackage{nicefrac}       
\usepackage{microtype}      
\usepackage{lipsum}		
\usepackage{doi}
\usepackage{algorithm}
\usepackage{algorithmic}
\usepackage{amsfonts,amsmath,amssymb,amsthm}
\usepackage{booktabs}
\usepackage{multirow}
\usepackage[capitalise]{cleveref}
\usepackage{textcomp}
\usepackage{xcolor}
\usepackage{comment}
\usepackage{subcaption}
\usepackage{multicol}
\usepackage{fixltx2e}
\usepackage{relsize}
\usepackage{mathbbol}
\usepackage{xurl}
\usepackage{wasysym}
\usepackage{bm}
\usepackage{graphicx}
\usepackage{xcolor}
\usepackage{hyperref}       
\usepackage{url}            
\usepackage{booktabs}       
\usepackage{amsfonts}       
\usepackage{nicefrac}       
\usepackage{microtype}      
\usepackage{graphicx}
\usepackage{xcolor}
\usepackage{markdown}
\usepackage{tcolorbox}
\usepackage{multirow}

\title{WildfireGPT: Tailored  Large Language Model for Wildfire Analysis}

\author{
    Yangxinyu Xie \thanks{Department of Statistics and Data Science, University of Pennsylvania, Philadelphia, PA. \texttt{\{xinyux, suw\}@wharton.upenn.edu}} \And Bowen Jiang \thanks{Department of Computer and Information Science, University of Pennsylvania, Philadelphia, PA. \texttt{\{bwjiang, cjtaylor\}@seas.upenn.edu}} \And 
    Tanwi Mallick \thanks{Mathematics and Computer Science Division, Argonne National Laboratory, Lemont, IL. \texttt{tmallick@anl.gov, rross@mcs.anl.gov}} \And
    Joshua David Bergerson
    \thanks{Decision and Infrastructure Sciences Division, Argonne National Laboratory, Lemont, IL. \texttt{\{jbergerson, jhutchison, dverner, jbranham, malexander, llevy\}@anl.gov}} \And
    John K. Hutchison $^{\S}$ \And
    Duane R. Verner $^{\S}$ \And
    Jordan Branham $^{\S}$ \And
    M. Ross Alexander $^{\S}$ \And
    Robert B. Ross $^{\ddagger}$ \And
    Yan Feng \thanks{Environmental Science Division, Argonne National Laboratory, Lemont, IL. \texttt{yfeng@anl.gov}} \And
    Leslie-Anne Levy $^{\S}$ \And
    Weijie Su $^{\ast}$ \And Camillo J. Taylor $^{\dagger}$
}

\hypersetup{
pdftitle={WildfireGPT: Tailored  Large Language Model for Wildfire Analysis},
pdfsubject={Large Language Models, Climate Change},
pdfauthor={Yangxinyu Xie, Tanwi Mallick},
pdfkeywords={Large Language Models, Climate Change}
}

\begin{document}

\maketitle

\input{sections/abstract}
\input{sections/introduction}

\input{sections/methodology}
\input{sections/experiment}
\input{sections/conclusion}

\section{Acknowledgement}
This material is based upon work supported by Laboratory Directed Research and Development (LDRD) funding from Argonne National Laboratory, provided by the Director, Office of Science, of the U.S. Department of Energy under Contract No. DEAC02-06CH11357.

\bibliographystyle{unsrt}
\bibliography{custom}

\input{sections/appendix}

\end{document}

%% file: sections/abstract.tex
\begin{abstract}
Recent advancement of large language models (LLMs) represents a transformational capability at the frontier of artificial intelligence. However, LLMs are generalized models, trained on extensive text corpus, and often struggle to provide context-specific information, particularly in areas requiring specialized knowledge, such as wildfire details within the broader context of climate change. For decision-makers focused on wildfire resilience and adaptation, it is crucial to obtain responses that are not only precise but also domain-specific. To that end, we developed WildfireGPT, a prototype LLM agent designed to transform user queries into actionable insights on wildfire risks. We enrich WildfireGPT by providing additional context, such as climate projections and scientific literature, to ensure its information is current, relevant, and scientifically accurate. 
This enables WildfireGPT to be an effective tool for delivering detailed, user-specific insights on wildfire risks to support a diverse set of end users, including but not limited to researchers and engineers, for making positive impact and decision making. 

\end{abstract}

%% file: sections/introduction.tex
\section{Introduction}

Understanding and adapting to climate change is paramount for professionals such as urban planners, emergency managers, and infrastructure operators, as it directly influences urban development, disaster response, and the maintenance of essential services. Nonetheless, this task presents a complex challenge that necessitates the integration of advanced technology and scientific insights. Large language models (LLMs) present an innovative solution, particularly in democratizing climate science. They possess the unique capability to interpret and explain technical aspects of climate change through conversations, making this crucial information accessible to people from all backgrounds \cite{rillig2023risks,bulian2023assessing, chen2023foundation, lu2024oxygenerator, atkins2024generative}. Recent research begins to augment LLMs with external tools and data sources to ensure that the information provided is scientifically accurate \cite{schick2023toolformer, qin2023toolllm, jiang2024multi}, such as ClimateWatch \cite{kraus2023enhancing} and findings from the IPCC AR6 reports \cite{vaghefi2023chatclimate}. 
Nonetheless, these existing studies primarily focus on one-off question-answering scenarios where a user asks one question and the LLM provides a direct answer. This approach, while effective for straightforward queries, {\it suffers from the assumption that a user comes with a fully formed, clear question}. The reality is more nuanced, especially in complex domains like climate change. Users may start with a vague idea or an unclear question, requiring several rounds of interactions to clarify their needs and thoughts. To that end, we prototype WildfireGPT, an LLM agent with a specific focus on wildfire analysis, designed to engage with the users actively and convert their wildfire-related inquiries into actionable insights. This is achieved through targeted questioning, strategic planning, and detailed analysis of a wide range of climate projections and wildfire-related literature. Moving beyond~\cite{vaghefi2023chatclimate, thulke2024climategpt, nguyen2024simulating}, WildfireGPT supports multi-round conversations that consider chat histories, addressing localized climate concerns based on asker's unique professional backgrounds and feedbacks. 
We aim to assist a wide range of wildfire hazard mitigation experts as end users in data analysis and policy development by providing comprehensive, reliable, and user-specific insights with data-driven evidence.

\begin{figure*}[t]
    \includegraphics[width=\textwidth]{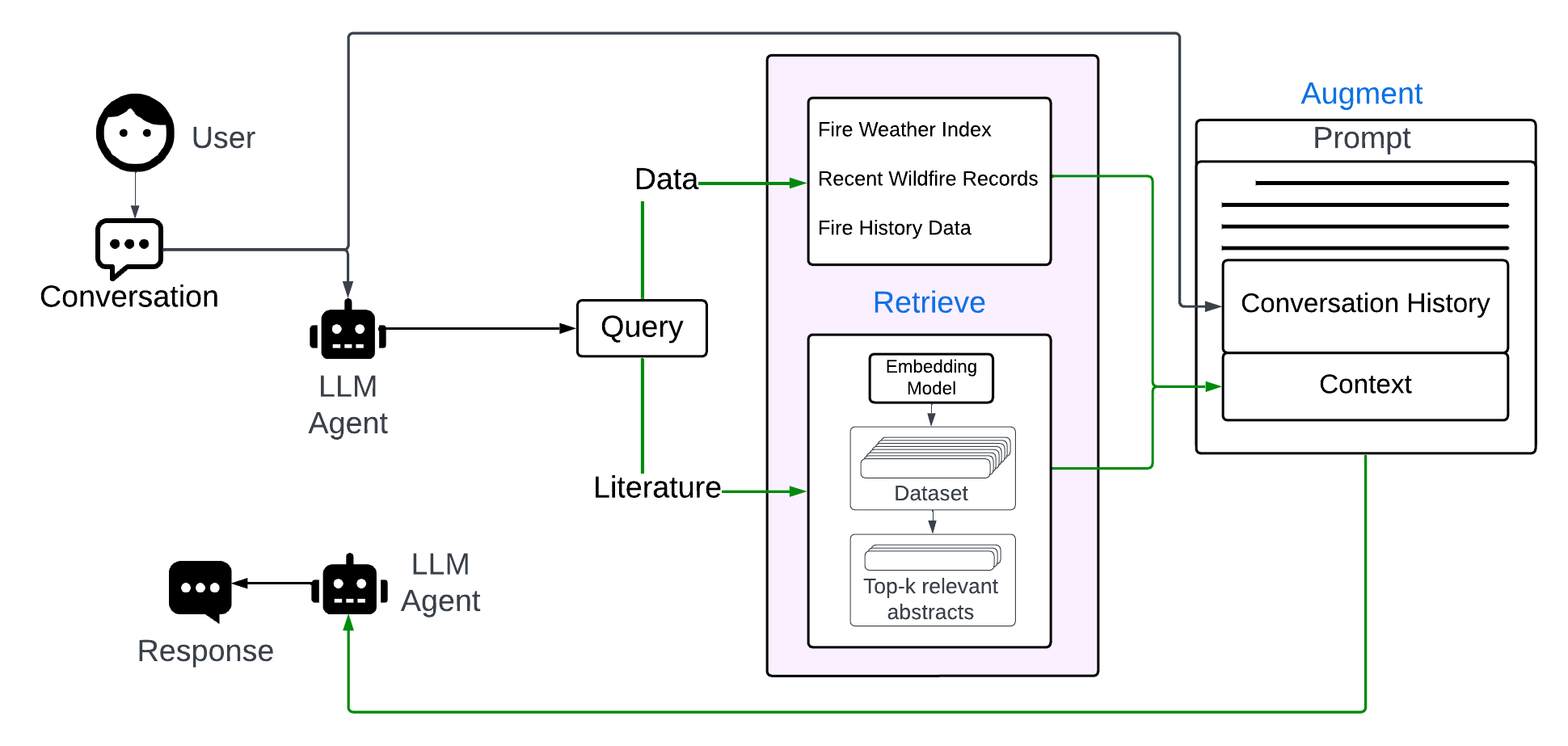}
    \caption{Here is an overview of the LLM agent integrated with the RAG framework. The LLM agent evaluates the user's input to determine the need for additional information. If needed, it retrieves climate projections and/or scientific papers. The agent then merges the retrieved data with its memory and a customized prompt to provide a response based on the augmented information.}
    \label{fig: rag}
\end{figure*}

%% file: sections/methodology.tex
\section{Methodology}

Our methodology comprises an LLM agent integrated with Retrieval-Augmented Generation (RAG) \cite{lewis2020retrieval, gao2023retrieval}, as illustrated in Figure \ref{fig: rag}. 
The agent transitions from a user profile module, a planning module to an analyst module: the agent first interacts with the user to create a profile and develop an action plan; then, the RAG enhances the LLM's response by providing additional information, such as climate projections and scientific literature along with the agent-generated context. Appendix~\ref{app: interface} demonstrates an example of the chat interface.


\textbf{Data Sources} WildfireGPT aims to provide comprehensive, user-specific insights on wildfire risks by leveraging diverse data sources. Location-specific data include Fire Weather Index (FWI) from Climate Risk \& Resilience Portal (ClimRR) \cite{climrr_anl}, recent wildfire records from the Wildland Fire Interagency Geospatial Services (WFIGS) Group \cite{nifc2024wildlandfire}, and the fire history data from the North American Tree-ring Fire Scar Synthesis (NAFSS) \cite{margolis2021noaawds}. 
To enhance users' understanding of wildfire-related climate change impacts on critical infrastructure, WildfireGPT also incorporates a corpus of scientific literature containing wildfire research articles from Community and Infrastructure Adaptation to Climate Change (CIACC) tool \cite{mallick2024understanding, mallick2024analyzing}.

\textbf{User Profile Module - Understanding the User's Concerns} The user profile module begins the conversation by asking the user a series of questions to gather information to establish a user profile surrounding the user's professional background, primary concerns, location of interest, urgency or timeline for addressing concerns, and specific aspects of wildfire risks they are interested in exploring. Throughout the process, the user is encouraged to provide detailed responses and can say, "I don't know," if they are unsure of how to respond to a question posed by the LLM. 

\textbf{Planning Module - Making and Communicating a Plan} In this module, an LLM is prompted to formulate a tailored action plan, outlining a sequence of steps to effectively address the user's queries. Typically, the plan has three parts: 1. Identify relevant datasets from the database described earlier. 2. Research academic papers. 3. Develop informed recommendations. This module aim to communicate with the user to ensure that the plan is aligned with the user's evolving requirements.

\textbf{Analyst Module - Analysis and Recommendations} Once the plan is formulated, the LLM agent transitions to the analyst module to execute the outlined action plan. At each step, the LLM agent addresses any follow-up questions from the users and asks them whether they would like to delve deeper into a specific topic or move on to the next step in the action plan. The analyst module is equipped with tools to retrieve and interpret data and scientific literature, following the RAG technique. Specifically, the system automatically decides whether to retrieve additional data based on the agreed-upon plan and the user's query. If no additional data is needed, the agent generates a response immediately. Otherwise, it considers two data types: \textit{1. location-specific data}, where the language model retrieves relevant tabular data and visualizations for the location identified by the user's query. \textit{2. literature review}, where the LLM agent generates a wildfire-related query based on the user's input, and employs a k-nearest neighbor approach with the all-MiniLM-L6-v2 embedder \cite{reimers2019sentence}, to extract the $k$ most similar manuscript abstracts from the scientific literature through Faiss vector store \cite{johnson2019billion}. Next, the language model receives the conversation history, along with the retrieved content, and a customized instruction to generate a response. This includes guidelines on interpreting summary statistics and citing the retrieved papers, ensuring a cohesive and informed output.

%% file: sections/experiment.tex
\section{Case Studies with Domain Experts} 
\begin{table*}[hbt!]
\centering
\footnotesize
\caption{Evaluation on correctness, relevance, entailment, and accessibility during each case study. The correctness scores are represented by precision, the number of correctly reported statistics divided by the total number of statistics reported by WildfireGPT. 
Other scores are in a format of (score, total times the domain expert opinion was collected to evaluate a response). Different case studies may require different number of evaluation points. The case studies show WildfireGPT performs well in correctness and relevance, with minor areas for improvement in entailment and accessibility.
}
\label{tab:correctness_scores}
\begin{tabular}{|p{0.25\linewidth}|c|c|c|c|}
\hline
Case Study & Correctness & Relevance & Entailment & Accessibility \\ \hline
Ecosystem Fire Management & 100.00\% (9/9) & 97.92\% (23.5/24) & 100.00\% (3/3) & 95.83\% (11.5/12) \\ \hline
Urban Wildfire Mitigation & 100.00\% (17/17) & 96.67\% (29/30) & 75.00\% (1.5/2) & 86.11\% (15.5/18) \\ \hline
Community Hazard Mitigation Planning & 90.91\% (10/11) & 100.00\% (46/46) & 100.00\% (6/6) & 100.00\% (24/24) \\ \hline
Post-Wildfire Public Safety & 100.00\% (4/4) & 98.21\% (27.5/28) & 100.00\% (3/3) & 100.00\% (15/15) \\ \hline
Overall Scores & 97.73\% (40/41) & 98.20\% (126/128) & 93.75\% (13.5/14) & 95.49\% (66/69)\\ \hline
\end{tabular}
\end{table*}

\begin{figure*}[hbt!]
\footnotesize
\centering
\begin{tcolorbox}[title=Case Study: Ecosystem Fire Management]
\textbf{Context:} WildfireGPT analyzed recent wildfire incident data and found that from 2018 to 2021, the months with recorded incidents were March (2 incidents) and July (1 incident), suggesting that March may be when controlled burns are typically implemented. To confirm this, WildfireGPT retrieved a study by \cite{stan2006dynamics}, which highlighted the importance of reintroducing fire in March for white oak regeneration in northeastern Illinois. Additionally, WildfireGPT analyzed seasonal Fire Weather Index (FWI) data, identifying lower risk periods during spring.
\vspace{1mm}
\\
\textbf{User Query:} Please suggest specific adaptations to fire management strategies to balance the utilization of fire for managing oak ecosystems and reducing the risk to residential and human developments.
\vspace{1mm}
\\
\textbf{WildfireGPT Response (Snapshot):}
\textit{Optimized Timing for Controlled Burns}:
\begin{itemize}
    \item Continue to utilize controlled burns in March when the FWI (Fire Weather Index) historically shows lower values, minimizing the risk of uncontrollable wildland fires. \vspace{-2mm}
    \item Schedule additional controlled burns during early spring or late autumn to take advantage of similarly low FWI scores, ensuring burns are conducted under optimal conditions for controlling spread and smoke dispersal.
\end{itemize}

\textbf{Expert Comment:} The answer effectively addresses a key concern: whether to continue conducting controlled burns in March. It confirms that March remains a suitable time for burns due to historically lower Fire Weather Index (FWI) values, which reduce the risk of uncontrolled fires.
The response also provides a nuanced suggestion by recommending additional burn periods in early spring and late autumn, taking advantage of similarly low FWI scores during these times, allowing for burns at both the beginning and end of the dormant season.
This strategy demonstrates a sophisticated understanding of fire management, as it balances the need for regular burns with the importance of conducting them under optimal weather conditions for safety and control. 
\end{tcolorbox}
\caption{Snapshot from the case study on ecosystem fire management: WildfireGPT demonstrates its ability to integrate data analysis and domain knowledge to provide actionable recommendations for ecosystem fire management. 
The domain expert's positive feedback highlights WildfireGPT's nuanced approach and its potential to support informed decision-making in wildfire management.}
\label{box:Ecosystem_Fire_Management}
\end{figure*}

To comprehensively evaluate WildfireGPT's capabilities, we conducted four in-depth case studies with domain experts specialized in critical infrastructure and fire mitigation, covering a diverse range of wildfire-related topics and geographical locations across the United States, including Ecosystem Fire Management in Naperville, IL, Urban Wildfire Mitigation in Beaverton, OR, Community Hazard Mitigation Planning in Mora Country, NM, and Post-Wildfire Public Safety in Sangre De Cristo Mountains, NM. We use GPT-4 \cite{achiam2023gpt} as our LLM with function calling capabilities and use Streamlit \cite{streamlit} for user-interface designs.

One notable example of WildfireGPT's effectiveness - in integrating data analysis with domain-specific knowledge - is demonstrated in the Ecosystem Fire Management case study. We show a snapshot in Figure~\ref{box:Ecosystem_Fire_Management}, including the feedback from the domain expert: a landscape manager sought guidance on adapting fire management strategies to balance the use of fire for maintaining oak ecosystems while minimizing risks to human development. WildfireGPT analyzed recent wildfire incident data, revealing that controlled burns were typically conducted in March. To validate this finding, the system retrieved relevant literature, including \cite{stan2006dynamics}, which highlighted the importance of reintroducing fire in March for promoting white oak regeneration in northeastern Illinois. Furthermore, WildfireGPT analyzed seasonal Fire Weather Index (FWI) data, identifying lower-risk periods during early spring and late autumn. Synthesizing these insights, WildfireGPT recommended optimizing the timing of controlled burns to coincide with historically low FWI values, thereby reducing the likelihood of uncontrolled fires. The domain expert highlighted WildfireGPT's nuanced approach, emphasizing its ability to balance ecological management goals with safety considerations. 

\textbf{Quantitative Evaluations}
To measure the correctness of WildfireGPT's responses, we adopt precision scores to quantify the proportion of correct statistics reported by the language model according to the retrieved information. 
Besides, we conducted expert evaluations during case studies, focusing on three key aspects to further assess the quality of WildfireGPT's responses: \textit{relevance, entailment, and accessibility.} Relevance assessed whether the model's responses is pertinent to the user's last question, profession, concerns, location, timeline, and scope. Entailment evaluates the logical coherence between the model's analyses/recommendations and the provided information. Accessibility examines the clarity and concision of the model's language, considering factors such as jargon, explanatory detail, and redundancy. Each question is rated on a three-point scale: "Yes" (1), "Could be better" (0.5), and "No" (0). We show results in Table~\ref{tab:correctness_scores} and leave detailed criteria in Appendix~\ref{app: criteria}. The table consistently shows high scores in correctness (97.73\%) and relevance (98.20\%). The entailment score is relatively lower partly due to fewer entailment questions in the case study.
While the accessibility scores are generally high (95.49\%), they show some variability, as different audience expects varying levels of specificity, suggesting the challenge of balancing detail and concision to meet diverse user needs.
Overall, the evaluation reflects strong performance, with few areas identified for improvement, such as WildfireGPT's occasional struggle to transition effectively between modules and a tendency for lengthy responses that make it challenging for experts to extract key information.

%% file: sections/conclusion.tex
\section{Conclusions and Future Work}
We introduce an LLM agent, WildfireGPT, representing a step forward in leveraging language models for specialized scientific domains such as climate change to promote positive impacts, with a particular focus on wildfire risks. This prototype shows great potential in transforming user queries into actionable insights, offering a customized interactive experience through strategic questioning and planning, and accessing climate projection data alongside scientific literature. Feedback from domain experts underscores the WildfireGPT's strengths in defining the scope of inquiries, establishing a clear roadmap for discussions, and enhancing productivity through efficient data retrieval and interpretation, which could be time-consuming for a user to accomplish independently. More case studies with different domain experts and climate data are currently underway to further showcase and validate WildfireGPT's capabilities in addressing real-world challenges.


%% file: sections/appendix.tex
\appendix
\newpage
\onecolumn
\section{Appendix}
\label{sec:appendix}

\subsection{Detailed evaluation criteria}
\label{app: criteria}

To further assess the quality of WildfireGPT's responses beyond statistical correctness, we conducted a human evaluation focusing on three key aspects: relevance, entailment, and accessibility. While the modular approach using precision scores quantifies the proportion of correct statistics reported by the language model, the human evaluation provides a more comprehensive assessment of the generated responses' usefulness and clarity for domain experts. For the evaluation, domain experts assessed the quality of the model's responses using a structured questionnaire. Each question was rated on a three-point scale, with "Yes" receiving a score of 1, "Could be better" a score of 0.5, and "No" a score of 0. 

\begin{table}[htbp]
\centering
\normalsize
\caption{Human evaluation criteria and corresponding questions used to assess WildfireGPT's performance in generating responses. Each question was rated on a three-point scale: "Yes", "Could be better", and "No".}
\label{tab:human_eval_criteria}
\begin{tabular}{|l|l|}
\hline
Criteria & Questions \\
\hline
\multirow{6}{*}{Relevance} & (1) Does my response answer your last question? \\
 & (2) Is my response relevant to your profession? \\
 & (3) Is my response relevant to your concern? \\
 & (4) Is my response relevant to your location? \\
 & (5) Is my response relevant to your timeline? \\
 & (6) Is my response relevant to your scope? \\
\hline
Entailment & (1) Do my analyses or recommendations logically follow from the \\
 & information (data, literature) provided? \\
\hline
\multirow{3}{*}{Accessibility} & (1) Does my response contain too many jargons? \\
 & (2) Does my response provide enough explanation? \\
 & (3) Does my response contain redundant or useless information? \\
\hline
\end{tabular}
\end{table}

\subsection{Data Sources and Retrieval Details of WildfireGPT}
\subsubsection{Quantitative Data Integration}

When the LLM provides a specific location after interacting with the user, we provide the following data sources to retrieve quantitative data relevant to climate change and wildfire risks:

\paragraph*{Fire Weather Index (FWI) Data Utilization} Sourced from the Climate Risk \& Resilience Portal (ClimRR) \cite{climrr_anl}, the FWI estimates wildfire danger using weather conditions that influence the spread of wildfires. The summary statistics includes historical average index (1995 - 2004), mid-century projections (2045 - 2054), and end-of-century projections (2085 - 2094). This temporal data enables the LLM to analyze and generate insights on the evolving nature of wildfire risks across different time periods.

\paragraph*{WFIGS Data for Recent Wildfire Records} The Wildland Fire Interagency Geospatial Services (WFIGS) Group provides comprehensive records of wildfire incidents \cite{nifc2024wildlandfire, irwin2024wildfireinfo}. Our LLM is configured to query and retrieve recent wildfire records from WFIGS, particularly focusing on yearly and monthly wildfire counts between 2015 and 2023, within 50 miles of the provided location. This retrieval capability allows the LLM to generate content that reflects recent trends and patterns in wildfire incidents, enhancing the model's relevance to current climate conditions.

\paragraph*{NAFSS and IMPD Data for Fire History Insights}: We also include metadata from the North American Tree-ring Fire Scar Synthesis (NAFSS) and the International Multiproxy Paleofire Database (IMPD) \cite{margolis2021noaawds}. When provided with a specific location, the LLM searches for records of fire-scarred trees within a 50-mile radius. If records are found, the LLM are provided with abstracts from the scientific studies on the three closest fire-scarred trees. This feature enables the LLM to provide detailed, location-specific insights into historical wildfire events, adding depth and precision to the generated content.

\subsubsection{Scientific Literature Retrieval}

To enhance our model's understanding of wildfire-related climate change impacts on critical infrastructure, we embedded abstracts of relevant scientific literature provided by Argonne National Laboratory \cite{mallick2023analyzing, mallick2024analyzing}. Utilizing SentenceTransformers \cite{reimers2019sentence}, these abstracts were transformed into embeddings, which were then stored in a Faiss vectorstore \cite{johnson2019billion}. For retrieval, we employed a $k$-nearest neighbor (kNN) approach, combined with the all-MiniLM-L6-v2 embedder, to identify and extract the most textually similar trajectories to the LLM agent's wildfire-related queries \cite{reimers2019sentence}.

\subsection{Chat Interface}\label{app: interface}
\begin{figure}[h]
    \centering
    \includegraphics[width=0.32\linewidth]{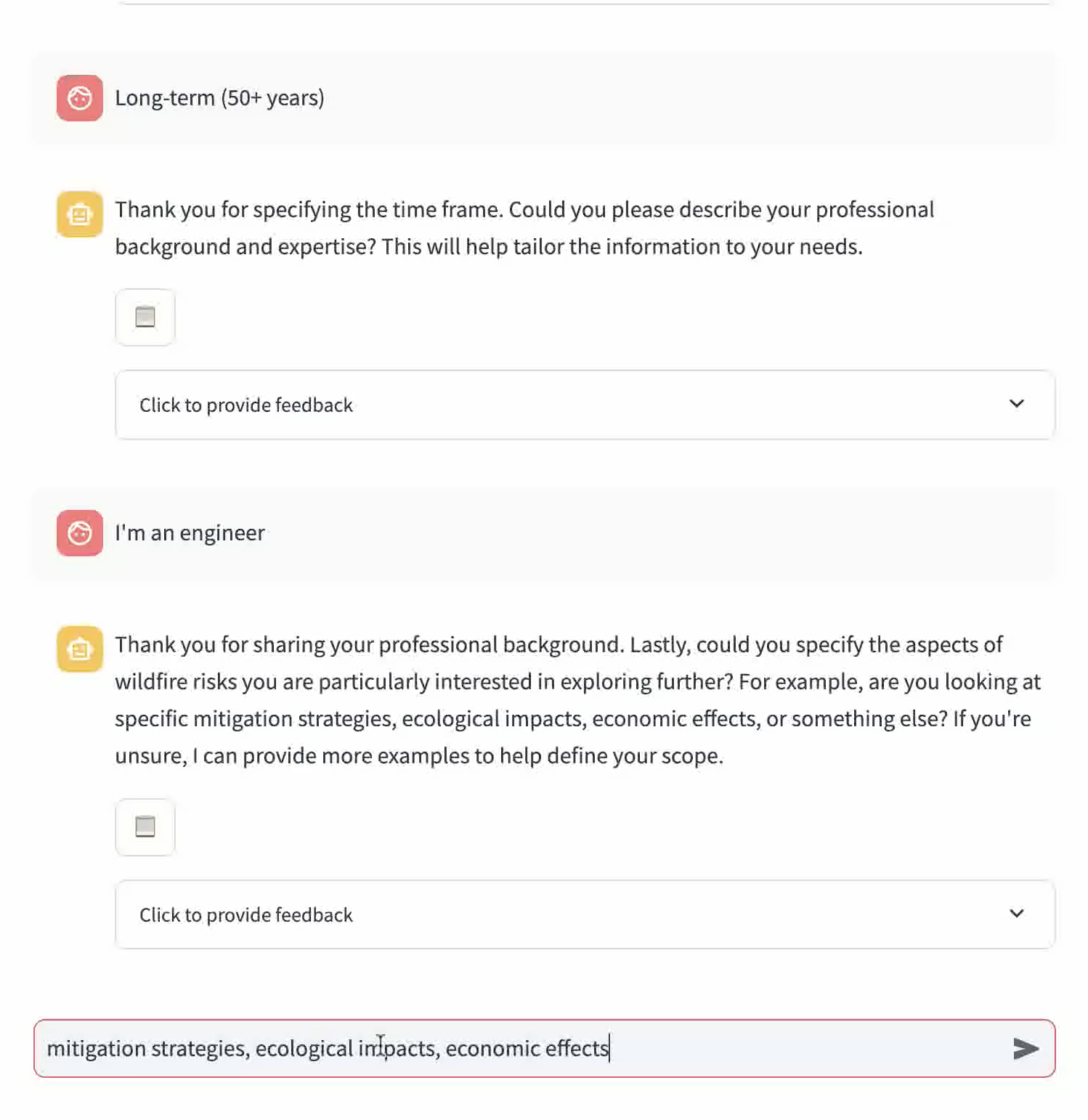}
    \includegraphics[width=0.32\linewidth]{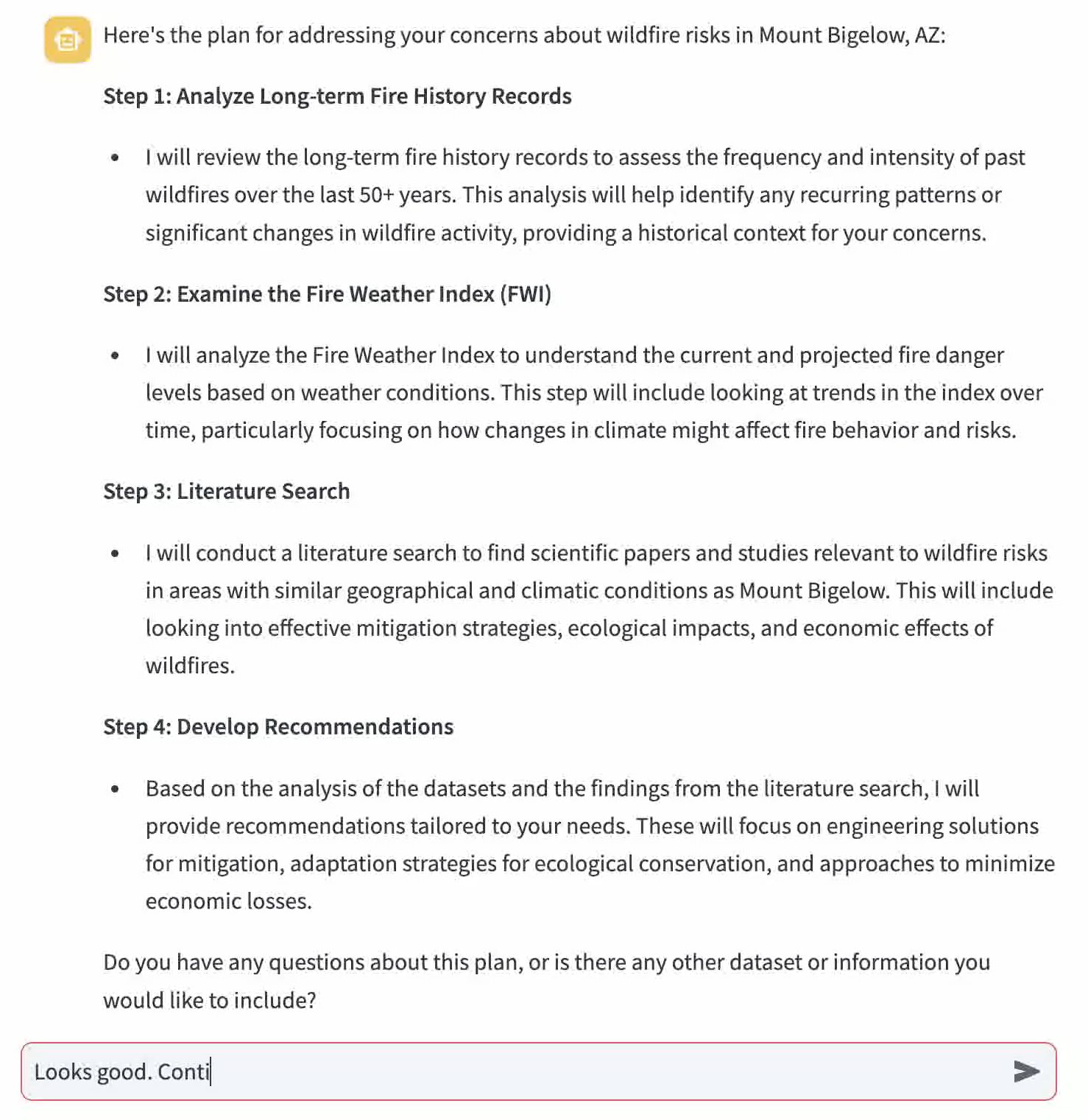}
    \includegraphics[width=0.32\linewidth]{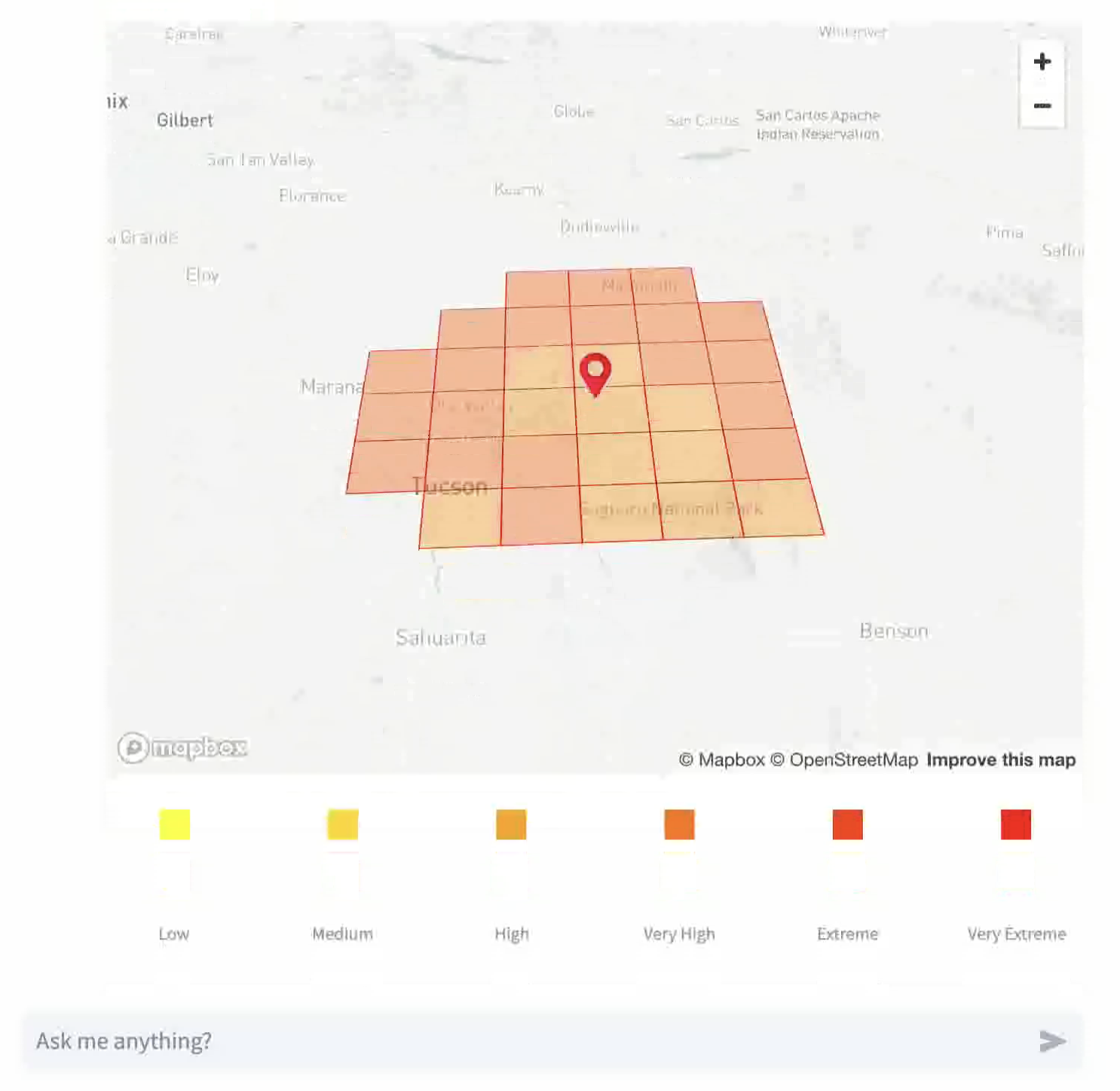}
    \includegraphics[width=0.32\linewidth]{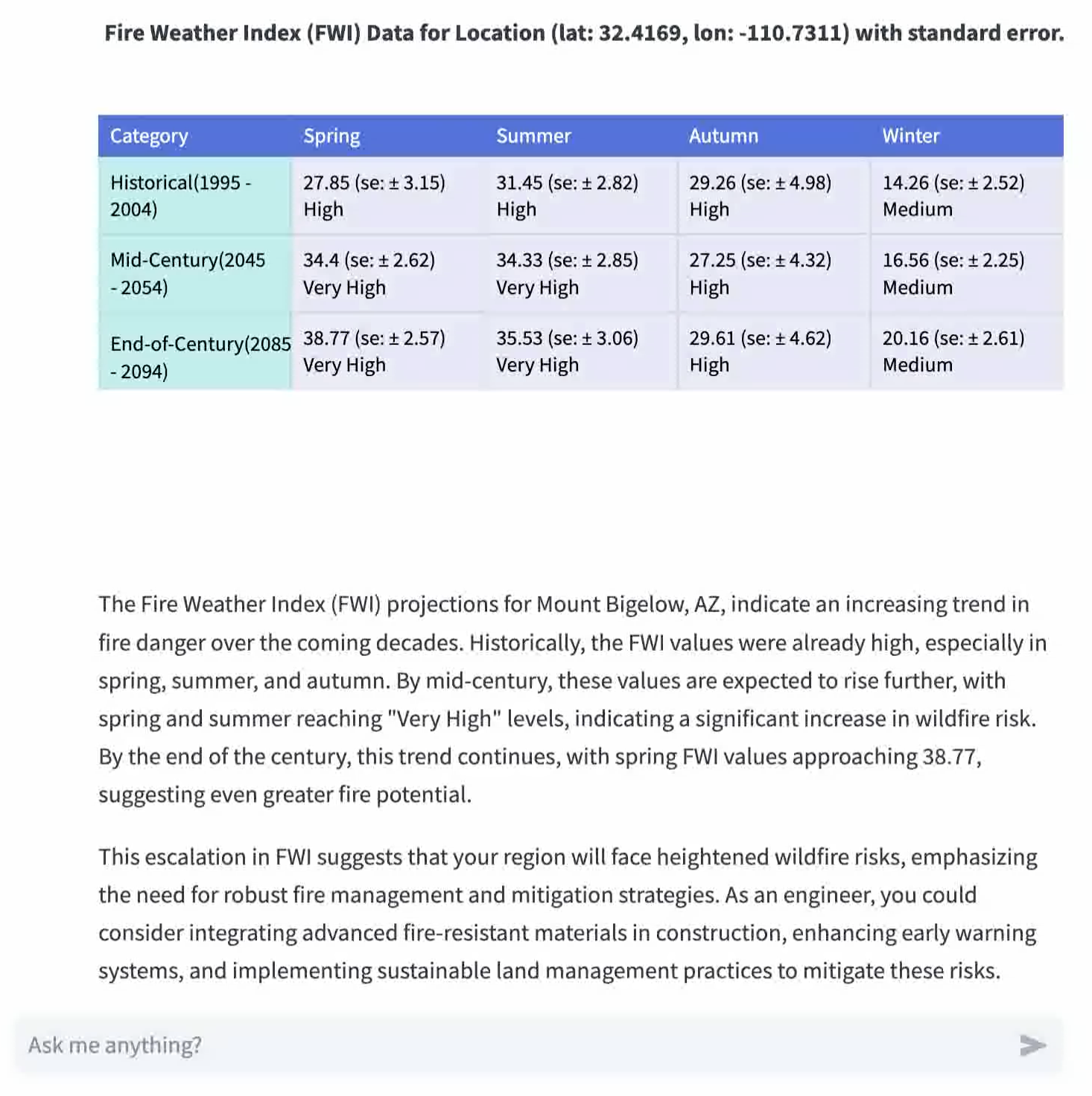}
    \includegraphics[width=0.32\linewidth]{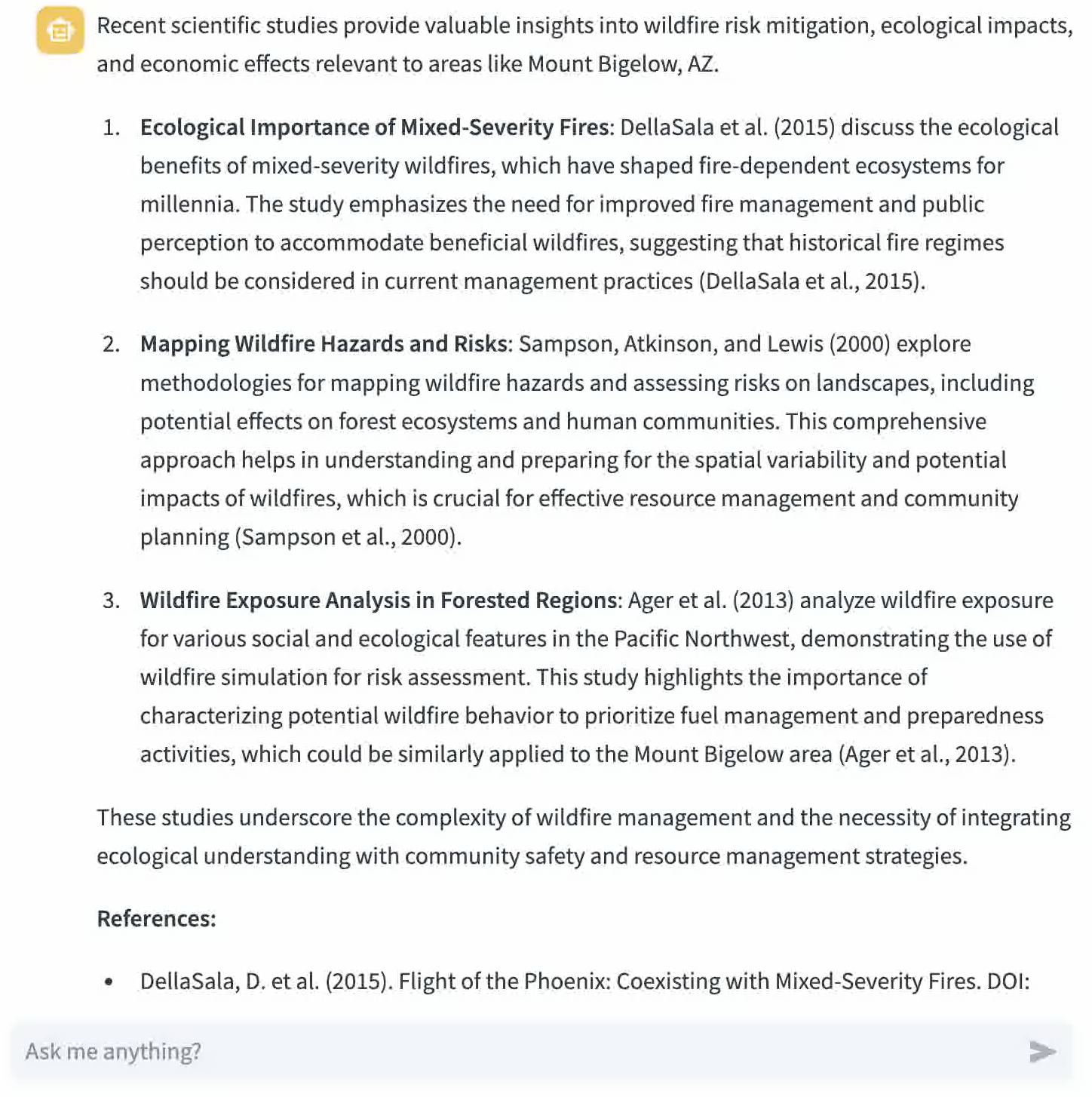}
    \includegraphics[width=0.32\linewidth]{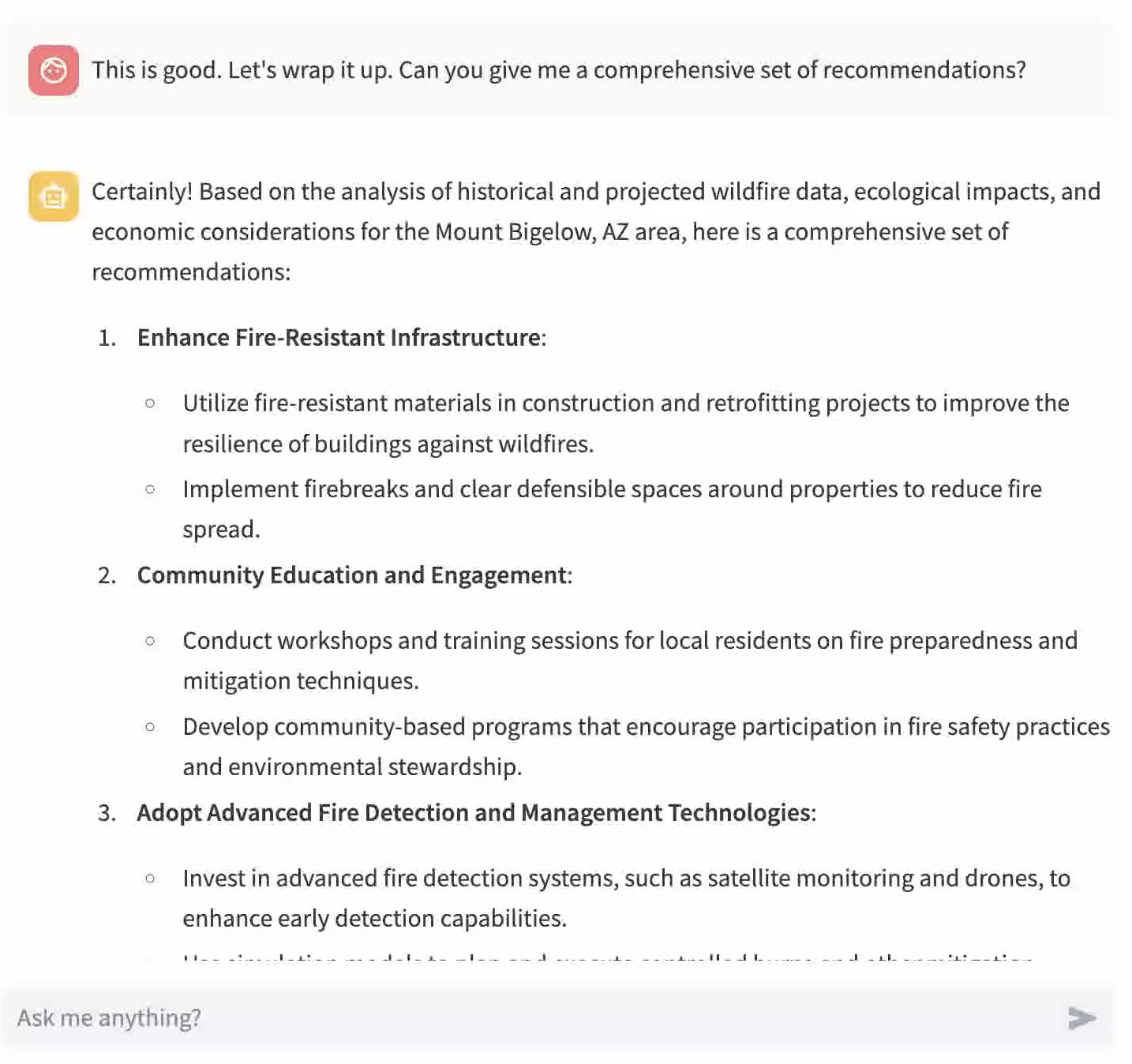}
    \caption{Overview of the WildfireGPT user experience. The screenshots are taken from one of the case studies themed comprehensive wildfire impact. The User Profile Module (top left) engages the user in a conversation to understand their background and concerns. The Planning Module (top middle) generates a tailored analysis plan based on the user's profile. The Analyst Module then executes the plan, analyzing Fire Weather Index data (top right and bottom left), conducting a literature review (bottom middle), and generating personalized recommendations (bottom right) to address the user's wildfire risk concerns.}
    \label{fig: interface}
\end{figure}




%% file: main.bbl
\begin{thebibliography}{10}

\bibitem{rillig2023risks}
Matthias~C Rillig, Marlene {\AA}gerstrand, Mohan Bi, Kenneth~A Gould, and Uli
  Sauerland.
\newblock Risks and benefits of large language models for the environment.
\newblock {\em Environmental Science \& Technology}, 57(9):3464--3466, 2023.

\bibitem{bulian2023assessing}
Jannis Bulian, Mike~S Sch{\"a}fer, Afra Amini, Heidi Lam, Massimiliano
  Ciaramita, Ben Gaiarin, Michelle~Chen Huebscher, Christian Buck, Niels Mede,
  Markus Leippold, et~al.
\newblock Assessing large language models on climate information.
\newblock {\em arXiv preprint arXiv:2310.02932}, 2023.

\bibitem{chen2023foundation}
Shengchao Chen, Guodong Long, Jing Jiang, Dikai Liu, and Chengqi Zhang.
\newblock Foundation models for weather and climate data understanding: A
  comprehensive survey.
\newblock {\em arXiv preprint arXiv:2312.03014}, 2023.

\bibitem{lu2024oxygenerator}
Bin Lu, Ze~Zhao, Luyu Han, Xiaoying Gan, Yuntao Zhou, Lei Zhou, Luoyi Fu,
  Xinbing Wang, Chenghu Zhou, and Jing Zhang.
\newblock Oxygenerator: Reconstructing global ocean deoxygenation over a
  century with deep learning.
\newblock In {\em Forty-first International Conference on Machine Learning},
  2024.

\bibitem{atkins2024generative}
Carmen Atkins, Gina Girgente, Manoochehr Shirzaei, and Junghwan Kim.
\newblock Generative ai tools can enhance climate literacy but must be checked
  for biases and inaccuracies.
\newblock {\em Communications Earth \& Environment}, 5(1):226, 2024.

\bibitem{schick2023toolformer}
Timo Schick, Jane Dwivedi-Yu, Roberto Dess{\`\i}, Roberta Raileanu, Maria
  Lomeli, Luke Zettlemoyer, Nicola Cancedda, and Thomas Scialom.
\newblock Toolformer: Language models can teach themselves to use tools.
\newblock {\em arXiv preprint arXiv:2302.04761}, 2023.

\bibitem{qin2023toolllm}
Yujia Qin, Shihao Liang, Yining Ye, Kunlun Zhu, Lan Yan, Yaxi Lu, Yankai Lin,
  Xin Cong, Xiangru Tang, Bill Qian, et~al.
\newblock Toolllm: Facilitating large language models to master 16000+
  real-world apis.
\newblock {\em arXiv preprint arXiv:2307.16789}, 2023.

\bibitem{jiang2024multi}
Bowen Jiang, Yangxinyu Xie, Xiaomeng Wang, Weijie~J Su, Camillo~J Taylor, and
  Tanwi Mallick.
\newblock Multi-modal and multi-agent systems meet rationality: A survey.
\newblock {\em arXiv preprint arXiv:2406.00252}, 2024.

\bibitem{kraus2023enhancing}
Mathias Kraus, Julia~Anna Bingler, Markus Leippold, Tobias Schimanski,
  Chiara~Colesanti Senni, Dominik Stammbach, Saeid~Ashraf Vaghefi, and Nicolas
  Webersinke.
\newblock Enhancing large language models with climate resources.
\newblock {\em arXiv preprint arXiv:2304.00116}, 2023.

\bibitem{vaghefi2023chatclimate}
Saeid~Ashraf Vaghefi, Dominik Stammbach, Veruska Muccione, Julia Bingler,
  Jingwei Ni, Mathias Kraus, Simon Allen, Chiara Colesanti-Senni, Tobias
  Wekhof, Tobias Schimanski, et~al.
\newblock Chatclimate: Grounding conversational ai in climate science.
\newblock {\em Communications Earth \& Environment}, 4(1):480, 2023.

\bibitem{thulke2024climategpt}
David Thulke, Yingbo Gao, Petrus Pelser, Rein Brune, Rricha Jalota, Floris Fok,
  Michael Ramos, Ian van Wyk, Abdallah Nasir, Hayden Goldstein, et~al.
\newblock Climategpt: Towards ai synthesizing interdisciplinary research on
  climate change.
\newblock {\em arXiv preprint arXiv:2401.09646}, 2024.

\bibitem{nguyen2024simulating}
Ha~Nguyen, Victoria Nguyen, Sar{\'\i}ah L{\'o}pez-Fierro, Sara Ludovise, and
  Rossella Santagata.
\newblock Simulating climate change discussion with large language models:
  Considerations for science communication at scale.
\newblock In {\em Proceedings of the Eleventh ACM Conference on Learning@
  Scale}, pages 28--38, 2024.

\bibitem{lewis2020retrieval}
Patrick Lewis, Ethan Perez, Aleksandra Piktus, Fabio Petroni, Vladimir
  Karpukhin, Naman Goyal, Heinrich K{\"u}ttler, Mike Lewis, Wen-tau Yih, Tim
  Rockt{\"a}schel, et~al.
\newblock Retrieval-augmented generation for knowledge-intensive nlp tasks.
\newblock {\em Advances in Neural Information Processing Systems},
  33:9459--9474, 2020.

\bibitem{gao2023retrieval}
Yunfan Gao, Yun Xiong, Xinyu Gao, Kangxiang Jia, Jinliu Pan, Yuxi Bi, Yi~Dai,
  Jiawei Sun, and Haofen Wang.
\newblock Retrieval-augmented generation for large language models: A survey.
\newblock {\em arXiv preprint arXiv:2312.10997}, 2023.

\bibitem{climrr_anl}
{Argonne National Laboratory}.
\newblock Climate risk \& resilience portal (climrr), 2022.
\newblock Accessed: 2024-01-21.

\bibitem{nifc2024wildlandfire}
National Interagency Fire~Center (NIFC).
\newblock Wildland fire incident locations.
\newblock NIFC Open Data.
\newblock Accessed 2024-01-21.

\bibitem{margolis2021noaawds}
E.Q. Margolis and C.H. Guiterman.
\newblock {NOAA/WDS Paleoclimatology - North American Tree-ring Fire Scar
  Synthesis (NAFSS)}.
\newblock NOAA National Centers for Environmental Information, 05 2021.
\newblock Accessed 2024-01-21.

\bibitem{mallick2024understanding}
Tanwi Mallick, Joshua~David Bergerson, Duane~R Verner, John~K Hutchison,
  Leslie-Anne Levy, and Prasanna Balaprakash.
\newblock Understanding the impact of climate change on critical infrastructure
  through nlp analysis of scientific literature.
\newblock {\em Sustainable and Resilient Infrastructure}, pages 1--18, 2024.

\bibitem{mallick2024analyzing}
Tanwi Mallick, John Murphy, Joshua~David Bergerson, Duane~R Verner, John~K
  Hutchison, and Leslie-Anne Levy.
\newblock Analyzing regional impacts of climate change using natural language
  processing techniques.
\newblock {\em arXiv preprint arXiv:2401.06817}, 2024.

\bibitem{reimers2019sentence}
Nils Reimers and Iryna Gurevych.
\newblock Sentence-bert: Sentence embeddings using siamese bert-networks.
\newblock {\em arXiv preprint arXiv:1908.10084}, 2019.

\bibitem{johnson2019billion}
Jeff Johnson, Matthijs Douze, and Herv{\'e} J{\'e}gou.
\newblock Billion-scale similarity search with gpus.
\newblock {\em IEEE Transactions on Big Data}, 7(3):535--547, 2019.

\bibitem{stan2006dynamics}
Amanda~B Stan, Lesley~S Rigg, and Linda~S Jones.
\newblock Dynamics of a managed oak woodland in northeastern illinois.
\newblock {\em Natural Areas Journal}, 26(2):187--197, 2006.

\bibitem{achiam2023gpt}
Josh Achiam, Steven Adler, Sandhini Agarwal, Lama Ahmad, Ilge Akkaya,
  Florencia~Leoni Aleman, Diogo Almeida, Janko Altenschmidt, Sam Altman,
  Shyamal Anadkat, et~al.
\newblock Gpt-4 technical report.
\newblock {\em arXiv preprint arXiv:2303.08774}, 2023.

\bibitem{streamlit}
{Streamlit}.
\newblock \url{https://www.streamlit.io/}.
\newblock Accessed: 2024-01-21.

\bibitem{irwin2024wildfireinfo}
{Wildfire.gov}.
\newblock Integrated reporting of wildfire information (irwin).
\newblock Wildfire.gov.
\newblock Accessed 2024-01-21.

\bibitem{mallick2023analyzing}
Tanwi Mallick, Joshua~David Bergerson, Duane~R Verner, John~K Hutchison,
  Leslie-Anne Levy, and Prasanna Balaprakash.
\newblock Understanding the impact of climate change on critical infrastructure
  through nlp analysis of scientific literature.
\newblock {\em Sustainable and Resilient Infrastructure}, pages 1--18, 2024.

\end{thebibliography}
